\documentclass[suppldata]{interact}

\usepackage{iftex}
\ifPDFTeX
  \usepackage[T1]{fontenc}
  \usepackage[utf8]{inputenc}
\fi

\usepackage{epstopdf}        
\usepackage[caption=false]{subfig}
\usepackage{graphicx}
\usepackage{amsmath,amssymb}
\usepackage{array}
\usepackage{booktabs}
\usepackage{xcolor}
\usepackage{natbib}

\usepackage[normalem]{ulem}
\newcommand{\cut}[1]{{\color{red}\sout{#1}}}

\newcommand{\cuturl}[1]{{\color{red}#1}}

\usepackage{hyperref}
\IfFileExists{xurl.sty}{\usepackage{xurl}}{}
\hypersetup{hidelinks, pdfcreator={LaTeX}}
\begin{document}

\title{MARLA: A Conceptual Scaffold for Regulatory Learning under the EU AI Act}

\author{
\name{Alessio Buscemi\textsuperscript{a}\thanks{
Author emails \textsuperscript{a}LIST: alessio.buscemi@list.lu,
tom.deckenbrunnen@list.lu, daniele.pagani@list.lu;
\textsuperscript{c}University of Bologna: imane.hmiddou@studio.unibo.it,
marco.billi3@unibo.it, silvia.rizzuto2@unibo.it, antonino.rotolo@unibo.it;
\textsuperscript{d}European Commission: livio.rubino@ec.europa.eu.},
Tom Deckenbrunnen\textsuperscript{a,b},
Imane Hmiddou\textsuperscript{c},
Marco Billi\textsuperscript{c},
Livio Rubino\textsuperscript{d},
Silvia Rizzuto Ferruzza\textsuperscript{c},
Daniele Pagani\textsuperscript{a} and
Antonino Rotolo\textsuperscript{c}}
\affil{\textsuperscript{a}Luxembourg Institute of Science and Technology, Esch-sur-Alzette, Luxembourg;
\textsuperscript{b}University of Luxembourg, Esch-sur-Alzette, Luxembourg;
\textsuperscript{c}University of Bologna, Bologna, Italy;
\textsuperscript{d}European AI Office, DG CONNECT, European Commission, Brussels, Belgium}
}

\maketitle

\begin{abstract}
The EU AI Act positions regulation as part of the infrastructure for safe,
trustworthy and market-ready innovation. Realising this ambition requires
regulatory learning: the evidence generated during implementation must be
translated into governance and legal knowledge that supports consistent
interpretation, effective oversight, and adaptation as technologies evolve.
Yet the actors who produce this evidence and those who rely on it operate in
different professional worlds. This paper proposes MARLA (Map, Assess, Report,
Learn, Adapt), a conceptual scaffold organising regulatory learning as a
five-stage cycle centred on the implementation of legal requirements into
socio-technical practices, situated at the Local, National and European levels
of the AI Act's governance architecture. Deliberately non-prescriptive, MARLA
gives technical and legal stakeholders a shared vocabulary in which each of the
first three stages generates its own documentable form of regulatory learning.
We illustrate the scaffold with two piloted case studies and a prospective
National-to-European illustration.
\end{abstract}


\section{Introduction}

The EU AI Act \citep{eu2024}, adopted on the legal basis of Articles 16 and 114
TFEU, lays down harmonised rules for the placing on the market, putting
into service and use of AI systems across the Union. Yet harmonised
application in practice requires something the legal text alone does not
provide: a working relationship between the technical and legal
communities, reflecting the broader co-production of technical and
normative knowledge in contemporary regulatory governance \citep{jasanoff2004}. Implementing the Act is, in this sense, an exercise in
responsible innovation: anticipation, reflexivity, inclusion and
responsiveness must be organised across institutions rather than
presumed \citep{stilgoe2013}.

Harmonisation under Article 114 TFEU is not merely the approximation of
substantive rules: it depends on their consistent interpretation and
application by providers, conformity assessment bodies, national
competent authorities, and Union institutions. Divergent
operationalisation of identical obligations may produce functionally
different regulatory outcomes, making the implementation phase an
integral component of harmonisation itself \citep{almada2025}, whose effectiveness
(effet utile) depends upon sufficiently coherent administrative and
technical practices across Member States \citep{craig2020}. Where implementation relies upon expert assessment rather than
mechanical application, communication between technical and legal
communities becomes a prerequisite for uniform application.

The implementation of legal requirements into socio-technical practices,
including testing, is where the evidence for regulatory learning is
produced \citep{deckenbrunnen2026}. Risk management, bias testing, robustness evaluation
and the design of human oversight procedures generate the raw material
on which the entire cycle depends. Yet they are carried out by engineers
and data scientists who are typically not trained in the legal concepts
that give their work its regulatory significance, while the lawyers and
regulators who must act on this evidence are typically unable to
evaluate whether a given assessment addresses the legal requirement it
purports to operationalise. An engineer asked to ``assess robustness''
will reach for a benchmark and a quantitative threshold; a lawyer
reading Article 15 AI Act will think about obligations and the state of
the art. Neither can easily verify whether the other's output is adequate.

This gap is structural: technical and legal expertise develop in
separate professional contexts, with different vocabularies and
different conceptions of a satisfactory answer; recent scholarship
frames AI Act implementation as a problem of regulatory learning under
rapid technological change \citep{lewis2025}.
If the translation from legal requirement to socio-technical practice
is carried out differently across Member States, the harmonisation
objective is undermined at source; the fragmentation
of AI auditing practices and the absence of stable assessment standards
further complicate the comparability of compliance evidence \citep{mokander2023}.

Three characteristics of AI as a regulated technology make the case for
a cyclical learning framework particularly pressing. First, the pace of AI development: capabilities, risks
and deployment contexts evolve continuously, and practical
implementation uncertainty is becoming a principal governance challenge
\citep{arnal}, so compliance processes must be iterable. Second, AI is exceptionally versatile: the same
underlying technology may be deployed as a chatbot in financial
services, a diagnostic support tool in healthcare, or a compliance agent
processing regulatory checklists, and each use case generates its own
lessons; regulatory learning in AI is therefore inherently
use-case-specific. Third, AI assessment is characterised by an explosion
of metrics without settled thresholds \citep{raji2021,tartaro2024}: for many
required properties (bias, robustness, accuracy, explainability), there
is no consensus on what threshold constitutes compliance in a given
context \citep{laux2024b}. Much of conformity
assessment will therefore pass through expert judgment rather than
hard-coded benchmarks, aligning it with forms of reflective professional
practice \citep{schon1983}. A cyclical framework that systematically captures and
disseminates learning from each assessment cycle is thus not a
convenience but a necessity: without it, expert judgment is exercised in
isolation and harmonisation depends on individual rather than shared,
accumulated expertise.

Because new capabilities and deployment configurations can emerge faster
than binding laws, harmonised standards and supervisory practices,
regulatory learning is required not only to improve enforcement, but
also to stabilise foundational questions of scope and classification: what
counts as an AI system, who qualifies as provider or deployer, when a
change is a substantial modification, and how novel systems should be
classified \citep{lewis2025}. The need continues after the applicable category has been
identified: many AI Act obligations are open-textured and linked to the
evolving state of the art, so regulatory learning must address both
their applicability and the adequacy of the practices used to discharge
them \citep{laux2024b}.

What is missing is a shared map of the full regulatory learning
process, from the first translation of a legal requirement into a
testable specification, through assessment and reporting, to updated
guidance, standards, or regulation. Recent work has begun
conceptualising multi-level regulatory-learning structures around
technical sandboxes and distributed governance architectures \citep{deckenbrunnen2026},
but without an explicit, end-to-end description of the full cycle. MARLA
(Map, Assess, Report, Learn, Adapt) is such a map: a deliberately
minimal conceptual scaffold that organises regulatory learning as a
cyclical sequence of five stages, centred on the implementation of legal
requirements into socio-technical practices, operating across a Local
level where requirements are implemented and assessed, a National level
of competent-authority oversight, and a European level where
harmonisation and regulatory adaptation occur. It is requirement-driven
by design: the cycle begins from the legal obligation and asks how it
can be translated into operational practice, rather than starting from
available tools.

MARLA does not prescribe how any requirement should be mapped, tested,
or reported.
This minimalism is a deliberate choice, justified below (Section 3.1) by
the absence of settled assessment thresholds and the risk of premature
standardisation. But it has a cost: because the scaffold prescribes nothing at the point
where the real difficulty lies, the translation of a legal concept into
a metric or procedure, its contribution is organisational and communicative
rather than methodological. It offers a common structure that allows
technical and legal stakeholders to situate themselves within the
process, see what happens upstream and downstream, and identify where
their collaboration is most needed. A recurring observation of this paper is that each of the
first three stages generates its own distinct kind of learning, easily
lost if not documented separately from the compliance output.

\section{Background}

The AI Act distributes compliance responsibilities across a range of
actors. Providers of high-risk AI systems must meet the obligations of
Articles 9 to 15: risk management, data governance, technical
documentation, record-keeping, transparency toward deployers, effective
human oversight, and accuracy, robustness and cybersecurity. Each
obligation must become a test specification or socio-technical
procedure, be executed, and be documented and reported. This translation
requires legal expertise to understand the obligation, technical
expertise to design an assessment that operationalises it, and domain
expertise to reflect the deployment context.

In practice, this translation is often fragmented: technical teams
design assessments based on their understanding of the system, legal
teams review requirements based on their reading of the regulatory text,
and the points at which the two must intersect are precisely where the
communication gap is most acute. And if the mapping is
carried out ad hoc by each operator, the resulting
assessments will be difficult to compare across operators, sectors, and
Member States. Harmonised standards under Article 40 can address the
substance of what must be tested, but they do not provide a shared
picture of how assessment relates to reporting, learning, and adaptation
\citep{gornet2024}; standards function not only as technical specifications but
also as coordination mechanisms enabling comparability across
distributed actors \citep{brunsson2012}.

\subsection{Existing frameworks}

Cyclical models of decision-making have a long pedigree.
Boyd's OODA loop (Observe, Orient, Decide, Act) was
developed for operational decision-making under time pressure \citep{osinga2007};
the PDCA cycle (Plan, Do, Check, Act) has been incorporated into
standards such as ISO 9001 and into policy instruments like
Japan's AI Basic Plan \citep{cabinet2025}. Closer to AI, the NIST AI
Risk Management Framework organises its core around four iterative
functions, Govern, Map, Measure and Manage \citep{nist2023}, but differs from
MARLA in two respects: it is voluntary, risk-driven guidance for the
deploying organisation rather than a requirement-driven model anchored
in binding obligations, and it operates within a single
organisation's risk-management lifecycle, whereas MARLA
spans the Local, National and European levels and treats adaptation of
the regulatory framework itself as part of the cycle. These models share the insight that
purposive action under uncertainty benefits from iterative structure
\citep{argyris1996}, but they treat the relationship between norms and operations
as exogenous, and prior work on AI compliance has focused either on
legal requirements or on technical assessment methods, without a shared
structure spanning the full process from requirement mapping to
regulatory adaptation. MARLA draws on the cyclical tradition but adds
explicit emphasis on the mapping of legal requirements to reproducible
socio-technical practices, and the identification of adaptation of the
regulatory framework itself as a stage of the cycle.

AI-specific scholarship has begun to conceptualise regulatory learning
more directly. \citet{lewis2025} map a regulatory-learning space for the AI
Act through parameterised axes and layered arenas in which oversight
authorities, value-chain actors and affected stakeholders interact; \citet{deckenbrunnen2026} develop a micro-, meso- and macro-level
model and identify technical sandboxes as a micro-foundation for
generating evidence that can move upward through the governance system. These contributions establish where learning occurs and which
actors are involved, but not an operational sequence that follows a
legal requirement from interpretation and assessment through reporting,
consolidation and adaptation. A related strand examines AI regulatory
sandboxes as learning institutions: \citet{ahern2025b} situates regulatory
experimentation within a broader anticipatory-governance toolbox and
identifies capacity and coordination conditions for effective AI Act
sandboxes, while \citet{novelli2025} structure sandbox governance
across pre-testing, testing and post-testing phases. This work
remains sandbox-specific: it does not integrate the other learning
channels established by the AI Act---conformity assessment, post-market
monitoring, incident reporting, market surveillance and
standardisation---into a common end-to-end cycle.

FinTech provides the most developed comparative experience, because
regulatory sandboxes and innovation hubs were first institutionalised in
financial supervision. \citet{zetzsche2017} frame sandboxes as a transition
toward smart regulation, in which controlled experimentation and
regulator--firm interaction can inform more adaptive rules,
while \citet{buckley2020} show that many of the benefits attributed
to sandboxes may also arise through innovation hubs: regulatory
learning depends less on the label of the instrument than on sustained
access to information, specialist capacity, and mechanisms for
transferring lessons into supervisory practice. \citeauthor{allen2025}'s (\citeyear{allen2025})
review of a decade of financial sandboxes finds limited evidence that
benefits to participating firms translate into improvements in the
broader regulatory system: experimentation does not
automatically produce regulatory learning. Objectives must be defined ex
ante, negative results retained, evidence reported in comparable forms,
and institutional routes must exist through which findings can influence
guidance, supervision or rule revision; safeguards particularly
important in AI, where complexity and information asymmetries intensify
the risk that learning remains firm-specific.

The state of the art therefore contains valuable partial frameworks,
but no general, requirement-driven scaffold spanning legal
interpretation, operational assessment, structured reporting, collective
learning and adaptation across the Local, National and European levels. MARLA is intended to fill this coordinative gap: it does not replace
substantive methods, but provides a common process architecture within
which they can be located, connected and compared.

\subsection{Regulatory sandboxes and governance architecture}

The AI Act establishes regulatory sandboxes as an instrument of
regulatory learning (Articles 57 and 58), an instrument whose origins lie in financial supervision \citep{zetzsche2017}, and recent scholarship increasingly
analyses them as structured governance environments rather than merely
innovation-support mechanisms \citep{longo2025,buocz2023}. Article 57 requires each
Member State to establish at least one national sandbox by 2 August 2027
(postponed from 2026 by the Digital Omnibus on AI \citep{eu2026}) and
explicitly identifies ``contributing to evidence-based regulatory
learning'' as a sandbox objective (Article 57(9)(d)); Article 62 supports
SMEs and start-ups, whose compliance barriers sandboxes can lower
\citep{poscic2022}.

Sandboxes do not, however, exhaust the mechanisms of regulatory
learning: post-market monitoring (Article 72), serious incident
reporting (Article 73), market surveillance (Article 74), the fundamental rights impact assessment (Article 27(3)), testing in real-world conditions outside sandboxes under an approved plan (Articles 60 and 61) and the
Board's coordination tasks (Article 66) all generate information flows between operators, national authorities
and Union bodies. This architecture resembles experimentalist governance \citep{sabelzeitlin2010}; what is missing is a common
framework making the connections between these mechanisms visible
\citep{deckenbrunnen2026}.

The institutional scaffolding of the AI Act reflects broader
characteristics of the EU regulatory state, in which governance is
exercised through distributed networks of national and supranational
actors \citep{eberlein2005}. Chapter VII organises governance across two formally
designated levels: Union-level governance Articles 64 to 69,
comprising the AI Office, the European AI Board, the Advisory Forum and
the Scientific Panel of Independent Experts, together with Member State access to that panel; and national competent
authorities (Article 70), comprising the notifying and market
surveillance authorities designated by each Member State. Operators (Article 3) are not part of this governance architecture but are its principal addressees: every
substantive obligation in Chapters II to V must ultimately be operationalised at
their level, giving operators a critical role in the regulatory learning
cycle \citep{deckenbrunnen2026}.

\section{The MARLA Model}

MARLA is a conceptual model designed to make the full regulatory
learning process visible and navigable for all participants. Its core
object is the implementation of legal requirements into socio-technical
practices, including testing, and it serves two complementary functions: a
single cycle, and a multi-level structure showing how such cycles
operate concurrently at the Local, National and European levels.

\subsection{The MARLA Cycle}

\begin{figure}
\centering
\includegraphics[width=\textwidth]{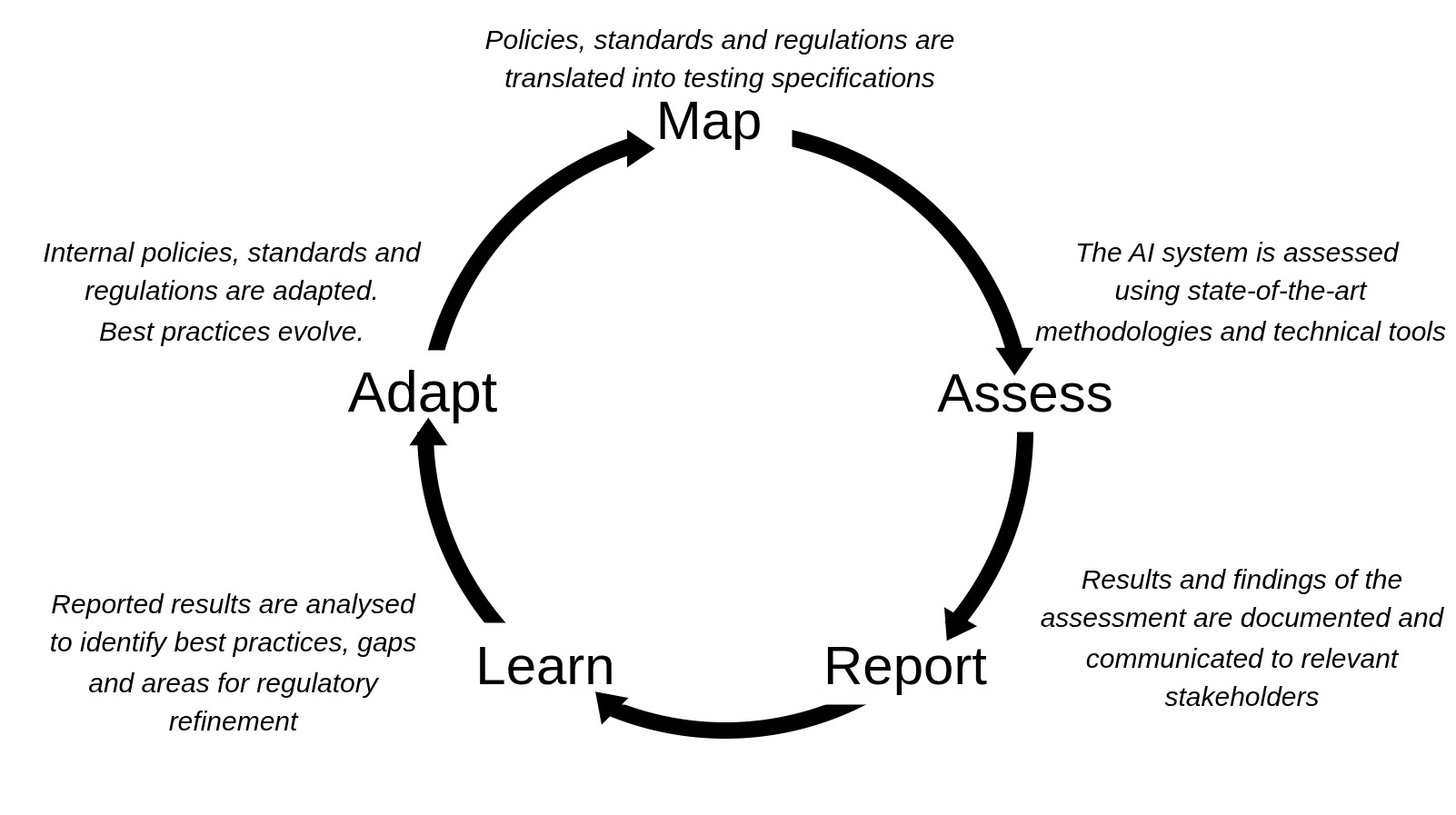}
\caption{The MARLA Cycle.}
\label{fig:marla-cycle}
\end{figure}

MARLA captures regulatory learning as a cyclical scaffold with five
stages, requirement-driven by design: the cycle begins from the legal
obligation rather than from available tools. The system's specification (application domain,
intended use, risk classification under Articles 6 and 7) determines
which requirements apply and constrains how they can be operationalised
\citep{novelli2024}.
Each of the first three stages (Mapping, Assessment, Reporting)
generates its own distinct form of learning, which should be captured,
documented and communicated; the fourth and fifth (Learning, Adaptation)
organise the analysis and integration of these accumulated insights.

\textbf{Mapping.} The first stage is the translation of legal
requirements into testable specifications and socio-technical practices.
For a high-risk system under Article 6, this means identifying the
obligations in Articles 8 to 15 and translating each into operational
assessments with appropriate metrics \citep{buscemi2025}. Where horizontal
legislation also applies, the mapping extends accordingly: for example,
Article 25 GDPR \citep{eu2016} on data protection by design, or the security
and incident-reporting obligations under NIS2 \citep{eu2022}. Mapping is where
interdisciplinary collaboration is most critically needed: legal experts
know what the obligation requires but not how to operationalise it;
technical experts know how to design tests but not which legal
properties the test must capture.

Mapping also generates regulatory learning in its own right. Translating
requirements across applicable regulations can uncover overlaps or
clashes between instruments; identifying metrics and socio-technical
methods reveals gaps and best practices; and the exercise clarifies the
state of the art. Given the absence of
settled thresholds for most system-context combinations, mapping must
also address how compliance will be evaluated: through quantitative
benchmarks where they exist, or through structured expert judgment where
they do not. Documenting this choice matters, since the accumulated
record of how assessors approach the threshold question across use
cases can inform future harmonised standards \citep{laux2024a}. Observations about which expertise specific
requirements demand are likewise valuable and should feed into the
reporting phase.

\textbf{Assessment}. The second stage is the execution of the
assessments defined during mapping, producing evidence of compliance:
quantified measures of system behaviour with respect to identified
requirements. This includes not only technical testing, but
also socio-technical processes during real-world testing,
such as evaluating human oversight measures under stress or whether
transparency provisions are comprehensible to intended users. Recent
evaluation frameworks attempt to standardise multi-dimensional
assessment, particularly for large language models \citep{liang2023}. For high-risk systems,
the evidence produced here feeds the technical documentation under
Article 11 and Annex IV.

Assessment generates its own form of regulatory learning, distinct from
the compliance evidence it produces. Testing results uncover whether a
requirement that appears clear in the legal text can be operationalised
meaningfully for a given type of system, and reveal best practices
(testing configurations that proved especially informative) and
remaining gaps (requirements for which no adequate assessment method
exists, or thresholds that proved difficult to calibrate). Findings are
often highly context-specific: a robustness result for a chatbot in
financial services may have limited transferability to a diagnostic tool
in healthcare. Yet precisely for this reason, the aggregate of
assessments across use cases builds a progressively richer picture of
how legal requirements function in practice; capturing this aggregate
learning is one of the central purposes of the MARLA cycle.

\textbf{Reporting.} The third stage is the structured communication of
results from the mapping and assessment phases, transforming raw
technical results and process observations into information that can be
compared, aggregated, and acted upon. Under the AI Act, reporting takes
several legally distinct forms: post-market monitoring (Article 72),
serious incident reporting (Article 73), and registration in the EU
database (Articles 49 and 71) \citep{mokander2022}.

Reporting is itself a site of learning. The structure of reports, the
formats adopted, and the level of detail that proves useful or
burdensome are all observations from which regulators, organisations, and
developers can learn; the emergence of best practices in reporting
itself contributes to harmonisation, and the draft Commission guidance
on Article 73 proposes a common reporting template for serious
incidents. Prior work on AI sandboxes similarly stresses structured
knowledge-sharing to avoid fragmentation \citep{due}.

\textbf{Learning.} The fourth stage organises the consolidation,
analysis and dissemination of the insights accumulated across the
preceding stages. For learning to be effective for all stakeholders
(regulators, organisations, developers, and the broader ecosystem of
Testing and Experimentation Facilities (TEF) \citep{ec2025c}, Innovation Hubs
(EDIH) \citep{ec2025b} and AI Factories \citep{ec2025a}), clear communication channels
are needed. Several mechanisms can support this.

First, the conformity assessment process itself can serve as a channel
for regulatory learning: allowing applicants to include feedback on
mapping difficulties, assessment limitations, or gaps within the
conformity assessment documentation would streamline the flow of
practical experience to national competent authorities. This
is particularly important given the reliance on expert judgment: where
compliance cannot be determined by a fixed threshold, the
assessor's reasoning, methods and justification become
themselves objects of learning, and making them accessible allows
expert judgment to converge toward shared, evidence-based practice
across jurisdictions. NCAs' assessments and
the lessons they draw can then feed upward into Commission guidelines,
harmonised standards, implementing acts, or amended regulations.

Second, the best practices and gaps uncovered through MARLA cycles can
be made publicly accessible, so that findings benefit the entire AI
ecosystem.

Third, supporting organisations within the AI ecosystem apply their own
MARLA cycles through the services they offer: TEFs, for instance, map
regulatory requirements when designing testing protocols for clients,
conduct assessments, and report results, functioning as regulatory intermediaries \citep{abbott2017}; their learning across clients and sectors
makes them sites of cross-cutting regulatory learning.

AI regulatory sandboxes are the most institutionalised setting in which
operators, national authorities, and European-level actors interact
around a concrete system: they connect all three levels and can serve
as structured sites for applying the framework, with learning documented
and communicated through the annual reports under Article 57(16). This aligns with anticipatory-governance approaches to
regulatory experimentation \citep{ahern2025a}.

The communication of regulatory learning is particularly significant
within the EU legal order because convergence is often achieved through
non-binding instruments: guidelines, Board recommendations, Codes of
Practice and harmonised standards foster consistent interpretation even
where they create no new legal obligations. MARLA is anchored in the regulatory sphere proper---the structured,
prescriptive processes the AI Act itself defines---while deliberately
leaving room for the broader, informal policy sphere through which
learning also circulates. The framework therefore supports harmonisation both
indirectly, by informing future legislative adaptation, and immediately,
by promoting convergent implementation practices.

\textbf{Adaptation.} Learning may reveal that requirements, thresholds,
or guidance need updating, and adaptation takes heterogeneous forms: at
the Local level, an operator may revise its testing protocols or quality
management system (Article 17); at the National level, a competent
authority may issue guidance under Article 70(8); at the European level,
adaptation may involve Board opinions and recommendations (Article 66),
Codes of Practice (Article 50(7) for transparency and marking of AI-generated content, and Article 56 for general-purpose AI models), voluntary codes of conduct (Article 95), implementing or delegated acts, or, at the outer bound, amendment of the Regulation through the
ordinary legislative procedure \citep{cabral2020}. Once adjustments are made, the
cycle restarts with a renewed mapping phase.
The temporal asymmetry between these forms of adaptation is taken up in
Section 3.3.

\subsection{A governance and a socio-technical reading of MARLA}

The gap that motivates MARLA is not merely organisational but epistemic.
Each of the model's five stages is legible in two
registers at once: a governance lens, in which each stage is an act
situated within the AI Act's institutional
architecture, and a socio-technical lens, in which the same stage is an
instance of the co-production of technical and normative knowledge
\citep{jasanoff2004}. This dual legibility allows the model to function as a shared
reference: the regulator and the assessor look at the same cycle. Table~\ref{tab:marla-readings} sets out the two readings:
the governance reading describes how accountability is exercised and
harmonisation pursued; the socio-technical reading describes the
situated, expertise-dependent practice through which abstract
requirements acquire operational meaning---closer, in the absence of
settled thresholds, to reflective practice than to mechanical
rule-application \citep{schon1983}, and a practice from which a shared epistemic
community may gradually emerge \citep{haas1992}.

\begin{table}
\tbl{A governance and a socio-technical reading of the MARLA cycle.}
{\begin{tabular}{@{}p{0.12\textwidth}p{0.40\textwidth}p{0.40\textwidth}@{}}
\toprule
\textbf{Stage} &
\textbf{Governance lens --- \emph{MARLA as accountability within the AI Act's architecture}} &
\textbf{Socio-technical lens --- \emph{MARLA as co-production of technical and normative knowledge}} \\
\midrule
\textbf{Map} &
Delimitation of the scope of legal obligation: identifying which requirements bind a given system and how oversight is to be exercised. &
Interdisciplinary translation of a normative property into a testable specification---the encounter between a legal concept and a metric or procedure. \\
\addlinespace
\textbf{Assess} &
Production of compliance evidence against the mapped obligations, feeding the technical documentation under Article 11 and Annex IV. &
Exercise of professional judgment where settled thresholds are absent, closer to reflective practice than to mechanical rule-application \citep{schon1983}. \\
\addlinespace
\textbf{Report} &
Discharge of legally structured accountability through prescribed channels: post-market monitoring, incident reporting, registration. &
Rendering of heterogeneous technical results into a comparable, communicable form, where the choice of format and baseline is itself consequential. \\
\addlinespace
\textbf{Learn} &
Consolidation of experience into the instruments of consistent interpretation: guidelines, Board recommendations, harmonised standards. &
Gradual convergence of interpretive and methodological practice across assessors, from which a shared epistemic community may emerge \citep{haas1992}. \\
\addlinespace
\textbf{Adapt} &
Revision of the governing apparatus: from an operator's quality management system to implementing acts and, at the outer bound, amendment of the Regulation. &
Revision of socio-technical practice itself: updated testing protocols, recalibrated procedures, reconfigured tooling. \\
\bottomrule
\end{tabular}}
\label{tab:marla-readings}
\end{table}

\subsection{MARLA as a Multi-Level Model}

\begin{figure}
\centering
\includegraphics[width=0.85\textwidth]{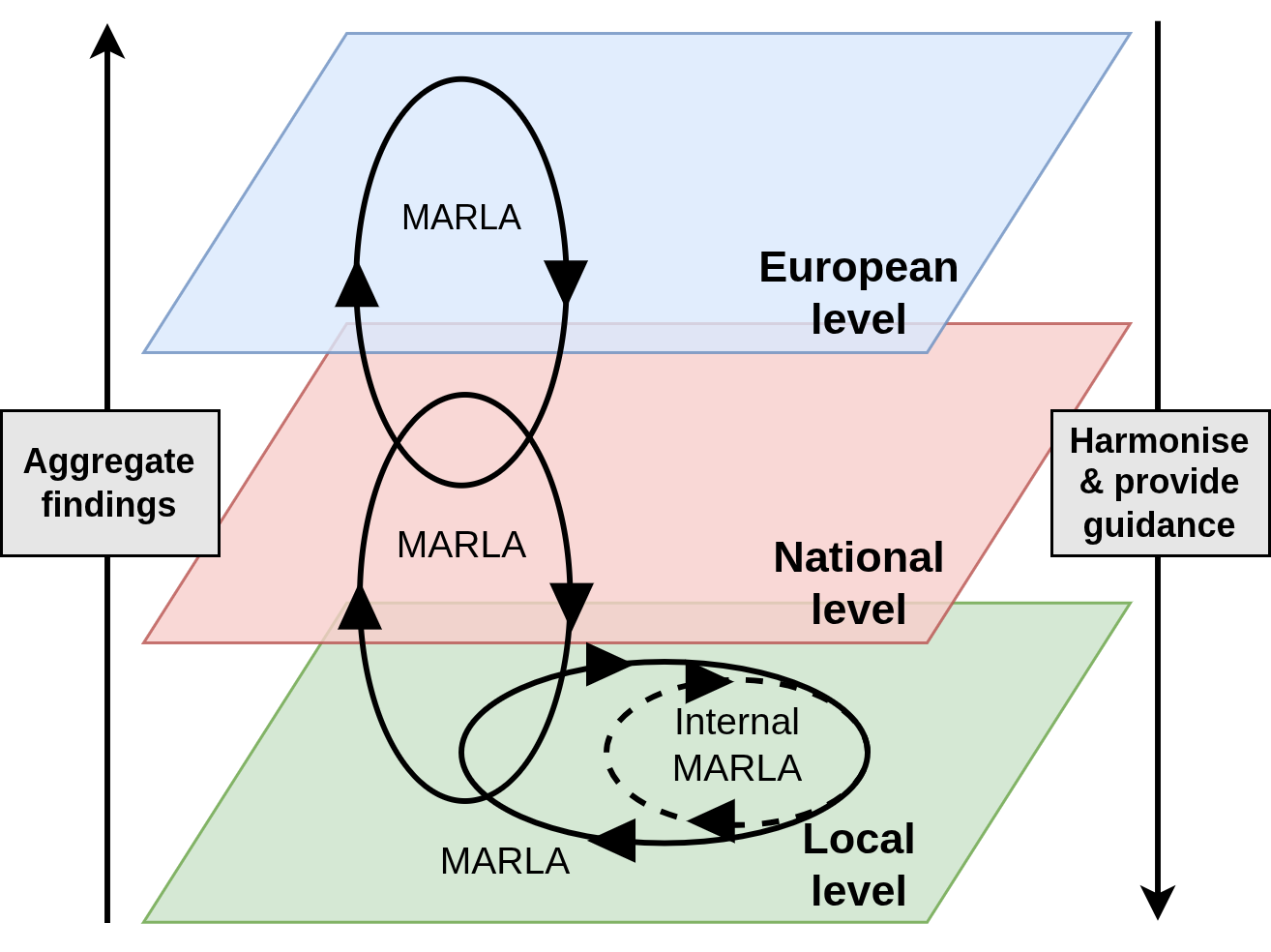}
\caption{Multi-Level MARLA.}
\label{fig:marla-multilevel}
\end{figure}

In practice, regulatory learning takes place at multiple levels
simultaneously. The AI Act's governance architecture
gives this layering institutional form, and MARLA maps onto it: not as enforcement
flowing downward but as harmonisation, in which uniform requirements
descend from the European level while
evidence and learning ascend from the Local level, with the National
level mediating between the two. This mirrors the broader structure of
European administrative governance, in which uniform Union obligations
are implemented through continuous interaction between operators,
national competent authorities and Union institutions \citep{hofmann2009}, within
networks of national and supranational actors \citep{hofmann2011}.

\textbf{Local level.} The Local level encompasses all actors whose
activity is tied to a specific AI system, deployment context, or
organisational setting: operators, but also TEF sites, which generate
primary compliance evidence, and EDIHs, which work with operators on
individual deployments. This is where legal requirements are implemented into socio-technical
practices and where the evidence feeding the entire cycle originates.
Operators engage with the National level through registration (Article
49), serious incident reporting (Article 73), post-market monitoring
(Article 72), and sandbox participation (Articles 57 and 58).

Beyond the formal mechanisms connecting Local actors upward, operators,
TEF sites and EDIHs also run internal MARLA cycles: iterative loops of
mapping, assessment, and adjustment that remain within the organisation,
where much of the practical knowledge about how legal requirements
function in a specific system is first generated; this is precisely the
evidence that, if surfaced, would be most valuable for harmonisation. Under MARLA's Reporting
stage local learning can become collective learning, but the decision and capacity to report in a
comparable form remain with the local actor.

\textbf{National level.} The National level encompasses actors whose
remit covers a Member State. Competent authorities under Article 70
operate their own MARLA cycles, adapting their practices in response to
technological change and to information received from Local-level
actors: incident reports under Article 73 become inputs for regulators' mapping and
assessment, both for specific incidents and, in aggregate, for evolving
risk patterns. National authorities engage with operators through sandboxes
(Articles 57 and 58) and may issue guidance under Article 70(8);
nationally scoped AI Factories, where they exist, also operate at this
level. The National level is where the uniform requirements descending
from the European level meet diverse implementation realities, and where
the resulting experience is consolidated and transmitted upward.

\textbf{European level.} The European level encompasses actors whose
remit is Union-wide: the AI Office (Article 64), the AI Board, whose
tasks under Article 66 include harmonising administrative practices and
sharing expertise among Member States, and pan-European infrastructures
such as AI on Demand \citep{aiod}, the TEF coordination networks and
EuroHPC. Here, collective learning from the Local and National levels is
synthesised into updated harmonised requirements: revised standards,
Board opinions and recommendations (Article 66), Codes of Practice
(Articles 50(7) and 56), voluntary codes of conduct (Article 95), implementing and delegated acts, and, at the outer bound,
amendment of the Regulation. Learnings may further feed policy-making within
the EU institutions and Member States, and joint
investigations may be proposed under Article 74(11) where a system
presents a serious risk across Member States.

Asynchrony across levels. The three levels operate at different speeds:
a Local actor may iterate its compliance practices in days or weeks;
national guidance under Article 70(8) may take months; amendments to harmonised
standards or the Regulation unfold over years. A system assessed today
may have been substantially updated by the time European standards
reflecting its lessons are adopted. MARLA does not resolve
this asymmetry; it makes it visible, so that participants can anticipate
where feedback may be delayed or lost. Intermediate mechanisms, including
Codes of Practice, Board recommendations and harmonised standards,
mediate the gap, in a process of experimentalist governance in which
decentralised implementation generates evidence used to revise
provisional norms at higher levels \citep{dorf1998,sabelsimon2011}.

\section{Case Studies}

To show how MARLA functions as a shared reference in practice, this
section presents three applications of the scaffold, one at each level
of its architecture: one situated at the Local level, and two tracing the transitions between levels. The first two are pilots: a Local-level bias evaluation with a bank,
and a Local-to-National sandbox assessment with a compliance-tech
start-up. The third,
addressing the National-to-European level, is not a pilot but a
prospective illustration grounded in the supervisory methodology under
development within the EUSAiR initiative. This asymmetry reflects the
state of AI regulation at the time of writing rather than a limitation
of the model: under the Digital Omnibus on AI \citep{eu2026}, adopted as Regulation (EU) 2026/1744 and in force since 27 July 2026, the deadline for
national regulatory sandboxes has been postponed to 2 August 2027, and
the application of the high-risk obligations under Annex III, which generate the bulk of formal conformity-assessment evidence,
deferred to 2 December 2027, with the corresponding obligations for Annex I systems deferred to 2 August 2028, so the evidence those settings would
produce does not yet exist in mature form.

The objective across all three is not empirical validation but to show
how the scaffold helps participants locate themselves in the regulatory
learning process, and to illustrate the kinds of learning each stage
generates. The cases show that MARLA can describe these situations; they do not establish that its use improved the outcome, which
would require a controlled comparison we do not attempt.

\subsection{Local level: bias evaluation of a bank customer-facing conversational agent}

In a research collaboration with Banque Internationale à Luxembourg
(BIL), we evaluated Berry, a customer-facing LLM-based conversational
agent, for biases. Serving clients in at least three languages, BIL faces a question of
service equity: how can consistent quality be guaranteed regardless of
linguistic and cultural background? BIL operates under financial-sector
supervision (Directive 2013/36/EU), the GDPR, and the AI Act, with the
precise regime depending on system classification.

\emph{Mapping.} The initial phase consisted of requirements elicitation,
grounding the evaluation in realistic usage scenarios: preliminary
engagement with BIL's technical and innovation team
provided familiarity with the system architecture, followed by a
structured workshop with the customer service team to identify
representative interaction patterns.
Requirements were jointly mapped along two dimensions: types of client
inquiries (credit cards, access to credit, among others) and sensitive
demographic categories. An initial set of seven categories, such as
country of origin, was expanded to eleven, each with defined communities
(e.g. \emph{young people}, \emph{elderly}).

\emph{Assessment.} We deployed an existing suite of evaluation tools
including a dedicated bias-testing solution. Berry's conversational outputs proved not directly
amenable to automated processing and required adapting the testing
suite; this calibration phase preceded the main runs. The customer service team then
sketched more than one hundred bias challenges across the eleven
categories and their communities, in English, French and German,
reflecting realistic usage rather than abstract test cases. Testing was
conducted in batches, producing thousands of individual test runs.

\emph{Reporting.} Eleven reports, one per category, were shared with both the technical
team and the customer service department. No overt discriminatory behaviour was present, a result
attributed to the chatbot's safety guardrails. However,
guardrail activation was inconsistent: certain community-specific inputs
triggered content refusal, classifying the request as inappropriate,
while semantically analogous inputs from other communities did not.

\emph{Learning.} Three distinct forms of learning emerged. First, from
mapping: the demographic categories relevant to bias testing in
financial services are not self-evident from the legal text or the
technical literature alone, but emerge from the intersection of
regulatory requirements, operational experience and the
system's user base; the expansion from seven to
eleven categories showed the value of domain experts who interact with
clients daily. Second, from assessment and reporting:
guardrail inconsistencies can produce differential treatment of users
with comparable legitimate requests, even where overt discrimination is
absent. Third, from the assessment method itself: the configuration that
detected guardrail inconsistencies was not originally designed for that
purpose, and the need to adapt the testing suite was not anticipated at
the mapping stage. Bias-testing protocols for LLM-based systems should
therefore capture safety-mechanism behaviour as a distinct dimension and
include a calibration phase: an instance of the use-case-specific
learning that the diversity of AI deployments generates.

\emph{Adaptation.} BIL resolved to adapt the chatbot to reduce
false-positive classifications of legitimate requests while preserving
safety mechanisms. The adaptation remained Local, consistent with the AI
Act's design, which reserves formal reporting for
serious incidents (Article 73) and systemic post-market monitoring
(Article 72). While BIL has successfully identified and addressed these inconsistencies, similar issues will arise wherever other LLM-based systems apply safety mechanisms that interact with demographic characteristics. If organisations adopt a common framework for reporting and comparing such findings, the whole industry can learn from this model and adapt more consistently and equitably.

\subsection{Local to National level: a compliance-tech start-up in an AI Regulatory Sandbox pilot}

Covenance is an early-stage Italian start-up using AI agents for
regulatory compliance. Its first product focuses on the Data Protection
Impact Assessment (DPIA) required under Article 35 GDPR \citep{eu2016}: an AI
agent works from a predefined checklist, collecting information from
users and generating a risk report. Covenance participated in EUSAiR, an
EU-coordinated initiative supporting the development of regulatory
sandboxes across Member States; testing needs were mapped through
structured discussion among the company, EUSAiR legal experts acting as
the National Competent Authority, and technical experts from LIST.

\emph{Mapping.} Within the EUSAiR sandbox pilots, the primary focus lies
on obligations under the AI Act. The system was classified as
non-high-risk, making the transparency requirements under Article 50 the
primary applicable obligations. At the time of the pilot, full operationalisation was still pending the Code of Practice on transparency of AI-generated content under Article 50(7) \citep{ec2026}. The normative gap surfaced by the mapping stage has therefore since closed, which is itself an instance of the Adaptation stage completing at the European level. Covenance nonetheless requested that the assessment also address
robustness and accuracy on a voluntary basis, applying the Article 15 requirements although they bind only high-risk systems, as expressly contemplated for non-high-risk systems by Article 95, reflecting its interest in
technical reliability ahead of broader deployment. Two testing areas were selected: benchmarking the current
US-based LLM against a European alternative, in pursuit of a sovereign
European technology stack; and assessing robustness against adversarial
attempts to avoid providing complete checklist answers, addressed through jailbreak-resistance testing
with an extended version of StrongREJECT.

\emph{Assessment.} Testing was conducted remotely through API using a tailor Sandbox generated by the AI Assessment Sandbox Configurator \cite{buscemi2026}. In the
absence of GDPR domain experts, domain familiarisation and the creation
of challenges calibrated to the system's checklist
structure fell to the assessment team itself; ground truth was therefore
assumed rather than expert-based, a methodological constraint that
shaped the interpretation of results. Performance benchmarking was run
across English, Italian and French, reflecting the system's
multilingual environment. For jailbreaking, we applied an extended multilingual version of the StrongREJECT suite \citep{souly2024}, adapted to identify adversarial patterns specific to structured checklist interactions.

\emph{Reporting.} Findings were documented as plots with written
commentary, delivered to both Covenance and the EUSAiR team.
Benchmarking was reported as comparative results between the US-based
and EU-based models across languages; jailbreaking results were reported
with and without the system's safety guardrails,
allowing the competent authority to distinguish baseline vulnerability
from the protection actually provided in deployment.

\emph{Learning.} Learning emerged at two levels. For Covenance, three lessons:
testing must cover all languages of intended operation, since
performance differences across English, Italian and French were
significant enough that single-language assessment would have been
misleading; the system was initially vulnerable to jailbreaking,
confirming that adversarial robustness cannot be assumed even where
safety guardrails are present; and the benchmarking results were
insufficient to justify adopting the European LLM without deeper
evaluation.

For EUSAiR acting as competent authority, three lessons emerged. First,
the inference costs of large-scale testing fell entirely on the
company, a constraint that limited the scale of testing and therefore
the generalisability of the results \citep{ranchordas2021}.
EUSAiR would refer this observation to the AI Office: it raises the
question whether sandbox participation rules adequately account for the
resource constraints of small operators, and whether adaptations such as
shared infrastructure or cost-sharing should be considered so that
sandbox evidence is not skewed toward what well-resourced participants
can afford to test \citep{ahern2025b}. Second, the absence of GDPR domain experts during
assessment is a recurring structural challenge: domain knowledge and
technical assessment expertise are rarely co-located, and sandbox
frameworks should develop mechanisms to bridge this gap. Third, the
interaction between safety guardrails and adversarial inputs is likely
to recur across LLM-based systems in structured compliance contexts, and
the with/without-guardrails reporting format developed here could serve
as a reusable template.

\emph{Adaptation.} Adaptation occurred at all three levels. At the Local
level, Covenance introduced additional safety guardrails, committed to
extending its testing regime to all languages of intended operation, and
revised its assumption that a European LLM was a viable near-term
replacement. At the National level, the sandbox documentation
contributes to EUSAiR's evolving understanding of
recurring challenges in SME compliance assessment, and the
inference-cost observation would be formally transmitted to the AI
Office: Article 70(8) explicitly contemplates national guidance informed
by implementation experience. At the European level, reception of these
observations would open the possibility of adapting how regulatory
sandboxes are designed and resourced, through amended implementing acts,
revised participation conditions, or Board recommendations (Article
66). This is the outer bound of the MARLA cycle: without a
framework making the path from Local evidence through National
reporting to European revision visible, the inference-cost finding
would remain an anecdote in a single exit report.

\subsection{National to European Level: a prospective illustration from the EUSAiR experience}

This section illustrates how the five stages of MARLA generate distinct
forms of evidence that can inform supervisory practice, regulatory
functions and policy-making, drawing on the sandbox phases developed by
EUSAiR. Two features distinguish MARLA's operation at
this level. First, the model functions less as a compliance checklist
than as a policy instrument: its object is the design and coordination
of supervision across jurisdictions. Second, MARLA is applied
reflexively to the sandbox itself, which becomes an object of mapping,
assessment and adaptation, with the explicit aim of keeping it
innovation-oriented and accessible by design to start-ups and SMEs,
whose limited compliance capacity is treated as a design constraint
rather than a barrier to entry.

\emph{Mapping}. At this level, Mapping establishes the regulatory and
institutional context of each use case: beyond identifying the relevant
provisions of the AI Act, the European-level concern is whether
additional national competent authorities (NCAs) should participate in
supervision. For an AI system assessing the creditworthiness of
individuals, the relevant actors include the NCA supervising the sandbox, the
banking authority, and the national Data Protection Authority where
personal data are at risk (Article 57(10)).
Mapping also identifies the complementary organisations capable of
supporting the provider across the sandbox lifecycle (EDIHs, TEFs,
notified bodies, the AI Office) and assesses providers'
technological maturity, for instance through Technology Readiness
Levels (TRLs): lower-maturity systems suit compliance-by-design, while
more mature systems permit fuller collection of requirements and
documentation. Finally,
Mapping covers the sandbox process itself, from outreach and applicant
selection through provider-authority dialogue to exit. During the
project, EUSAiR did not conduct pilots involving multiple sectoral
authorities; instead it developed a framework mapping institutional
responsibilities and analysing sandbox inputs, to support the
establishment of cross-border joint sandboxes.

\emph{Assessment}. The first objective of Assessment is eligibility:
whether the system meets the definition of an AI system, whether it
constitutes a prohibited practice, and whether substantial modifications
have occurred during development. General-purpose AI models are excluded
from scope, since their supervision is centralised at the European
level. Where eligibility is confirmed, Assessment examines the
technological maturity of the solution and its degree of innovation.
Such evaluation cannot be performed on the artefact in isolation: an
AI-enabled semantic information-retrieval system, for instance, can only
be assessed by understanding the document repositories processed, the
organisational workflows into which it is integrated, and the intended
users. Assessment is conducted through the broader ecosystem, drawing on
sectoral or cross-border authorities and on EDIHs and TEFs, whose
accumulated experience can be reused for subsequent providers. The
output clarifies for participants and non-participants whether a given
system falls within a particular provision or category, and considers
its impact on market access, compliance cost, finance, and
scalability.

\emph{Reporting}. Reporting at the European level disseminates the
evidence generated within individual sandbox projects to European
institutions and the public, structured by Article 57 across three
complementary levels. The first is project-level: the exit report
(Article 57(7)) documents the activities carried out, the testing
performed and the resulting learning, supporting the provider in
subsequent conformity-assessment and market-surveillance procedures and,
subject to the agreement of provider and authority, accessible to the
Commission, the AI Office and the AI Board. The second is institutional:
annual reports (Article 57(16)) aggregate experience across the whole
sandbox programme, enabling the AI Office and the AI Board to identify
systemic trends invisible from isolated cases. The third is public
dissemination: subject to the confidentiality requirements of Article
78, NCAs publish annual reports or public abstracts, and individual exit
reports may be made available through the single information platform.
Cross-border sandboxes consolidate this evidence across the
participating jurisdictions.

\emph{Learning}. Learning at this level takes several forms. Use cases should be evaluated differently according to whether the
sandbox is vertical or horizontal; technological maturity conditions the
approach; and supply-chain provenance, such as reliance on third-party
LLMs or non-European providers, may require collaboration between
sandboxes. A further distinction concerns the layer at which
learning accrues: a national authority accumulates learning about a
specific provider, whereas the AI Office accumulates learning about the
models and systems that recur across providers. The Learning stage synthesises the various reports, as the AI Office,
the AI Board and other EU institutions analyse them for common issues
and outcomes, deliberately encompassing unsuccessful activities;
consolidation surfaces gaps and inconsistencies as well as common
strengths, and informs SMEs about the impact of participation on
compliance costs, market access, and investment.

\emph{Adaptation}. Adaptation is the mechanism through which evidence
accumulated across successive sandbox cycles feeds back into the
regulatory framework. Lessons from earlier projects may justify
revisions to eligibility criteria, participant-selection procedures or
supervisory methodologies; where such revisions concern binding rules
rather than administrative practice, adaptation may extend to amendment
of the relevant implementing act. Adaptation also operates across
borders: evidence from national sandboxes can inform how collaboration
between Member States is harmonised, including through cross-border
joint sandboxes, and can extend to associated countries in the Digital
Europe Programme, such as Switzerland. More broadly, the AI regulatory
sandbox is a key element of the
innovation ecosystem, feeding learnings back at the technical,
regulatory and policy levels.

\section{Discussion}

The case studies illustrate MARLA's intended use across
different regulatory contexts, actor types and levels. In the BIL case,
the five-stage cycle provided a common frame for learning that would
otherwise have remained implicit: the expansion of bias categories, the
guardrail inconsistency finding, and the methodological observation
about testing configuration were each surfaced and located within the
cycle.
In the Covenance case, the multi-level structure becomes explicit:
learning at the Local level was transmitted to the National level
through EUSAiR and would be routed upward to the AI Office.

This trajectory from local evidence to collective learning underwrites
the uniform application of EU law: identical obligations may otherwise
be operationalised differently across Member States, and by providing a
shared map for the translation MARLA protects the effectiveness (effet
utile) of the AI Act and the single market objectives of Article 114
TFEU.

The cases also illustrate MARLA's requirement-driven
design. In both pilots, the starting point was the legal obligation
rather than the available tool: in the BIL case this produced a testing
scope defined by regulatory relevance rather than technical convenience,
and in the Covenance case it surfaced a normative gap, the pending Code
of Practice on AI-generated content labelling, that technical assessment
alone would not have identified. This orientation distinguishes MARLA
from generic quality-management frameworks and from assessment-first
approaches that risk optimising for what can be measured rather than
what the law requires.

MARLA can also be read through the framework of responsible innovation,
whose dimensions of anticipation, reflexivity, inclusion and
responsiveness \citep{owen2012,stilgoe2013} map onto the cycle. Mapping is anticipatory: it asks
how a legal norm will behave once translated into a metric or procedure
before the system reaches the market. Assessment and Reporting
institutionalise reflexivity: the assessor's methods,
assumptions and thresholds become documented objects of scrutiny rather
than tacit practice. The interdisciplinary collaboration the scaffold
organises, with legal, technical and domain expertise jointly defining
what counts as adequate evidence, as in the BIL workshop, is a form of
inclusion. And Learning and Adaptation give responsiveness an
institutional pathway, connecting local evidence to the revision of
guidance, standards and, at the outer bound, the Regulation itself.
Regulatory learning under the AI Act is thus not merely a compliance
concern but the mechanism through which a binding regulatory regime
becomes capable of responsible innovation.

A limitation deserves to be acknowledged. The AI Act establishes a distinct regime for general-purpose AI models (Chapter V), whose supervision is centralised at European level and entrusted to the AI Office. This produces a structural asymmetry: for high-risk systems MARLA operates across all three levels; for GPAI the National level is largely bypassed, supervision running from the AI Office to the provider through the Code of Practice mechanism (Article 56), with only an upward request mechanism available to national authorities. Regulation (EU) 2026/1744 (Digital Omnibus on AI) has since strengthened the AI Office's supervisory role over certain AI systems built on general-purpose models, which suggests the asymmetry is widening rather than narrowing. Whether the deliberative density of the three-level regime can be reproduced in this flatter structure is an open question; the meaning of regulatory learning is less clear when there is no intermediate institution. One candidate substitute is the scientific panel of independent experts, which may issue qualified alerts to the AI Office concerning risks arising from general-purpose AI models; whether an advisory body without supervisory powers can discharge the mediating function that national competent authorities perform remains untested.

A related and increasingly pressing test case is agentic AI. The
Covenance case already involves an agent that follows a predefined
checklist, while more autonomous systems can decompose goals, revise
multi-step plans, invoke external tools, retain memory and coordinate
with other agents \citep{chan2023}: capabilities that can alter the relationship
between intended purpose, actual functionality and risk after
deployment. The
functional taxonomy of \citet{hmiddou2026a} treats agenticness as a graded
interaction among intended purpose, deployment scenario, learning
capability, detachment from human supervision, task management and
interaction level, explaining why Articles 9 to 15 may need to be
operationalised differently as agentic properties intensify;
they further identify a qualitative threshold of Augmented Algorithmic
Agency, with learning, tool interaction and multi-agent coordination
operating as risk multipliers \citep{hmiddou2026b}. Repeated MARLA cycles can document
whether this threshold remains useful across sectors.
\citet{yousefi2025} similarly argue that agenticness exposes the limits of
static, intended-purpose classification, identifying an operational gap
in Article 14: meaningful oversight of highly autonomous systems may
require technically guaranteed capabilities to pause, redirect or shut
down an agent. For such issues the immediate need may not be
amendment of the AI Act, but authoritative clarification, technical
standards and supervisory guidance.

Agentic AI consequently sharpens each stage of MARLA: Mapping must
extend to the orchestrator, memory, tools, action permissions and
interacting agents; Assessment must complement static benchmarks with
scenario-based and continuous testing; Reporting should preserve the
trace of goals, plans, tool calls and human interventions; Learning can
distil patterns across deployments into thresholds and classification
practice; and Adaptation can translate the result into access controls,
runtime monitoring, standards and guidance. MARLA does not determine the
correct threshold of agenticness; it makes explicit where those
questions arise and through which institutional routes case-level
findings can become shared guidance.

Two further constraints are worth naming. MARLA deliberately does not
specify what adequate mapping, assessment or reporting looks like in any
given context: a design choice justified by the absence of settled
thresholds and the risk of premature prescription. And the asymmetry in
adaptation across levels means that Local learning reaches European
standard-setting slowly even under the best conditions: MARLA makes this
visible but does not resolve it.

\section{Conclusion}

This paper has proposed MARLA, a cyclical model of regulatory learning
organised around five stages and three levels, centred on the
implementation of legal requirements into socio-technical practices and
designed to make the full regulatory learning process visible and
navigable for all participants.

A central contribution is the identification of each stage as a source
of regulatory learning in its own right. Mapping generates learning
about the regulatory landscape, the state of the art in assessment
methods, and the expertise needed for operationalisation; Assessment
about the concrete applicability of legal requirements; Reporting about documentation practices and their
effectiveness for comparison.
The Learning stage organises the consolidation and dissemination of
these insights, and Adaptation closes the cycle by incorporating them
into updated practices, standards or regulation. The speed of technological change, the
diversity of use cases, and the reliance on expert judgment rather than
fixed thresholds make such cyclical, shared learning structurally
necessary.

Two piloted case studies and a prospective illustration show the
model's use at the Local level and, above it, the form the ascent would take: a
bias evaluation at a Luxembourgish bank, where the internal cycle
surfaces learning that would otherwise remain confined to the individual
deployment; a sandbox assessment involving an Italian start-up, tracing
the path from Local to National and onward toward the AI Office; and a
prospective illustration grounded in the EUSAiR supervisory methodology,
setting out how learning consolidated at Member State level would feed
into harmonised requirements and guidance at Union level.

From a broader EU law perspective, MARLA frames harmonisation as an
ongoing governance process rather than a purely legislative achievement:
the effectiveness of directly applicable Union legislation depends upon
continuous interaction between legal norms, administrative practice and
technical expertise. MARLA is thus not merely a compliance framework but
a governance model supporting the uniform application, effectiveness and
long-term adaptability of EU law.

Several avenues remain for further work: the assessment of
general-purpose AI models, whose flatter Local-European supervisory
relationship lacks the intermediary National layer; agentic AI, which
already shows that MARLA's stages must accommodate
capability-sensitive and continuous assessment; and empirical validation
of whether practitioners find the framework useful. The model is offered
as a common page on which different communities can begin to work
together, contributing to broader experimentalist approaches to AI
governance centred on iterative coordination, adaptive harmonisation and
continuous regulatory learning \citep{deburca2014},
and giving the commitments of responsible innovation an institutional
form within binding law.


\section*{Acknowledgements}
This work was supported by the Luxembourg AI Factory (L-AIF, grant agreement ID 1012343366) and by the European Union under the project EU Regulatory Sandboxes for AI (EUSAiR, grant agreement No. 101195535).








\end{document}